\documentclass[11pt,a4paper]{article}
\usepackage[hyperref]{acl2020}
\usepackage{times}
\usepackage{latexsym}

\usepackage{microtype}

\usepackage{graphicx}
\usepackage{amsmath}
\usepackage{amsfonts}
\usepackage{mathrsfs}
\usepackage[switch]{lineno}
\usepackage{multirow}

\aclfinalcopy

\title{Multi-Domain Dialogue State Tracking based on State Graph}

\author{Yan Zeng \\
  DIRO, Université de Montréal \\
  \texttt{yan.zeng@umontreal.ca} \\\And
  Jian-Yun Nie \\
  DIRO, Université de Montréal \\
  \texttt{nie@iro.umontreal.ca} \\}

\date{}

\begin{document}
\maketitle
\begin{abstract}
We investigate the problem of multi-domain Dialogue State Tracking (DST) with open vocabulary, which aims to extract the state from the dialogue. Existing approaches usually concatenate previous dialogue state with dialogue history as the input to a bi-directional Transformer encoder. They rely on the self-attention mechanism of Transformer to connect tokens in them. However, attention may be paid to spurious connections, leading to wrong inference. In this paper, we propose to construct a dialogue state graph in which domains, slots and values from the previous dialogue state are connected properly. Through training, the graph node and edge embeddings can encode co-occurrence relations between domain-domain, slot-slot and domain-slot, reflecting the strong transition paths in general dialogue. The state graph, encoded with relational-GCN, is fused into the Transformer encoder. Experimental results show that our approach achieves a new state of the art on the task while remaining efficient. It outperforms existing open-vocabulary DST approaches. 
\end{abstract}

\section{Introduction}

\begin{figure}[t]
\centering
\includegraphics[height=3.6in]{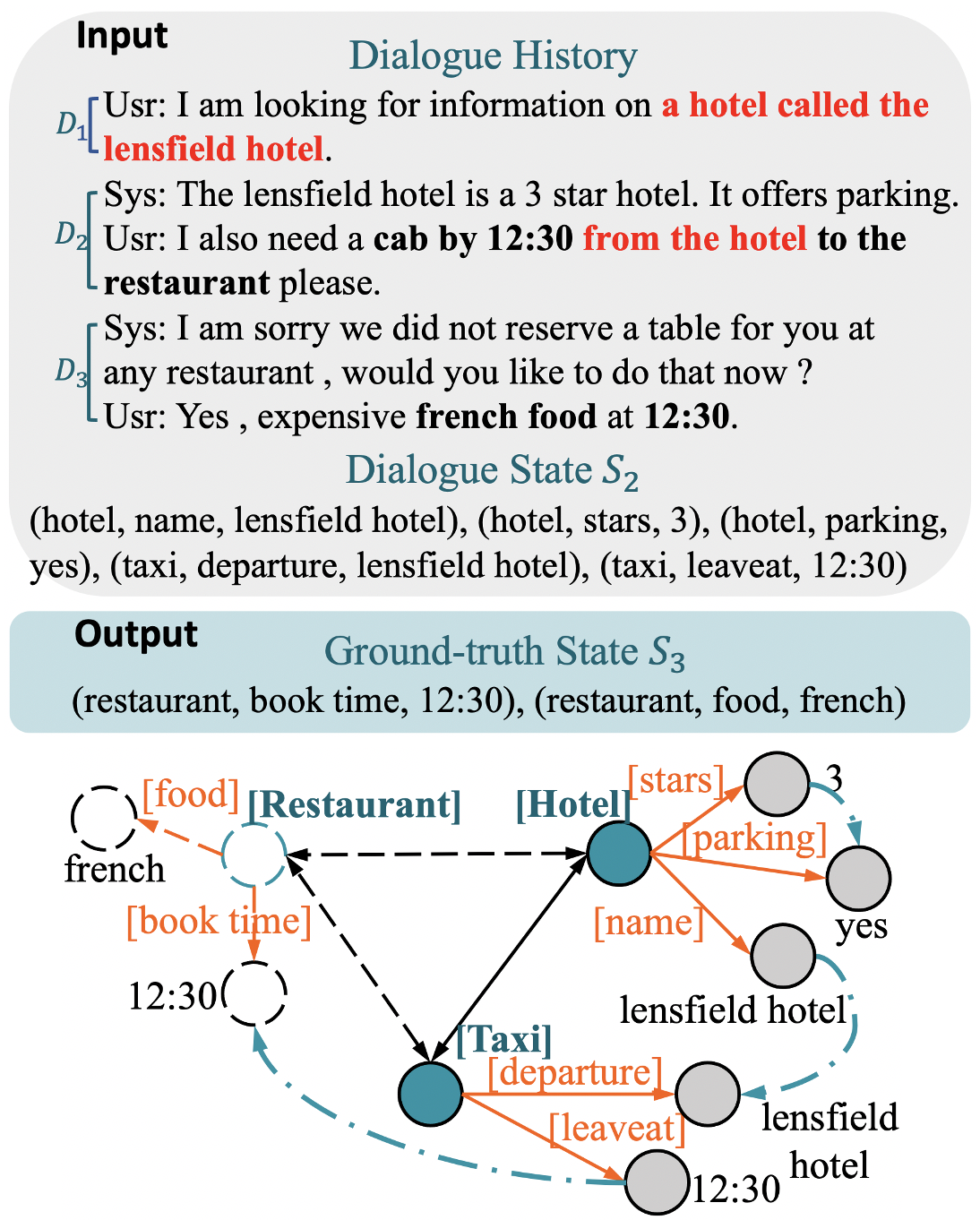}
\caption{An example of multi-domain DST and the corresponding dialogue state graph. Dash circles and lines are not included in the graph, but need to be predicted at this turn. Green dot dash arrows show how inference on the state graph helps DST.}
\label{Fig:GDST_intro}
\end{figure}

Dialogue state tracking (DST) is a core component in task-oriented dialogue systems. Accurate DST performance is crucial for appropriate dialogue management, where user intention determines the next system action. Recently, \citet{budzianowski2018multiwoz} introduced a new challenge -- DST in mixed-domain conversations. A user can start a conversation by asking to book a hotel, then book a taxi, and finally reserve a restaurant. Given \textit{dialogue history} and \textit{dialog state}, a DST model decides which (\textit{domain}, \textit{slot}) pairs to update values and extracts the \textit{values} from the input to fill them. A dialog state is described by a set of (\textit{domain}, \textit{slot}, \textit{value}) tuples. Figure \ref{Fig:GDST_intro} shows an example, where the goal is to extract the output state from the dialogue history and the previous state.

Traditionally, ontology-based DST is used, which assumes that not only the domain-slot structure has been defined, but also all possible \textit{values} for slots are known. The assumption largely simplifies DST into a classification/ranking task \cite{henderson2014word, mrkvsic2016neural, zhong2018global}. However, in realistic applications, it is often difficult to define the possible values for slots in advance\cite{xu2018end,wu2019transferable}. Thus, recent studies have focused on the open-vocabulary setting, where \textit{values} are not predefined and need to be directly extracted (generated) from the input. 
As the example in Figure \ref{Fig:GDST_intro} shows, in the current turn, the user needs a taxi departure from the hotel, and a DST model needs to refer to the hotel name that is expressed in the first dialog turn. To do this, earlier work  \cite{wu2019transferable, le2020non} encodes the full dialogue history as model input. Recent work \cite{kim2019efficient, zhu2020efficient} shows that the previous dialogue state can already serve as an explicit, compact, and informative representation of the dialogue history; thus we do not need to use the full dialogue history, but only a partial dialogue history, to combine with it. In general, the concatenated input is fed into a bi-directional Transformer for encoding, based on which, new state is extracted. 

It is interesting to observe that the attention-based approaches have produced very good results. The bi-directional self-attention based encoder could be viewed as a graph neural network whose input is a huge fully-connected graph on tokens -- any two tokens (i.e. graph nodes) in the input, whether they are in dialogue history or in dialogue state, can be connected via attention. In other words, the encoder learns to make inference by observing how tokens are connected in the 
large graph. This observation indicates that the underlying graph structure can be very useful for DST. Indeed, given a graph in which the tokens of hotel ``name" and taxi ``departure" are present, attention can be established between them, meaning that we can infer the taxi departure place from the hotel name.
However, the underlying graph used by transformer contains a large number of nodes, and any pair of nodes can be connected through attention. It may often happen that attention connects two nodes that are not relevant to the dialogue, thus leading to wrong inferences.

We believe that a better DST should start with a better graph, representing the key entities (domains, slots, and values) occurring in the dialogue history, as well as strong connections among them. In particular, the strong connections do not only rely on the dialogue history, but also on the general ways to conduct dialogues. For example, the fact that hotel booking is often evoked in human conversations (training data) when a train is booked means that the latter topic often naturally leads to the former (a natural transition), thus a strong connection should be created in the graph.
Such a better structured graph can encode complementary and more global information to the token-level graph created from the given dialogue history and input state, and help boost the accuracy of DST. 
Therefore, in this work, we build a new graph to represent (domain, slot, value) tuples 
as shown in Figure \ref{Fig:GDST_intro}. In addition, a co-occurrence edge is created to connect two domain nodes if they co-occur in the dialogue state (e.g. between \textit{Hotel} and \textit{Taxi}). The more such co-occurrences occur in the training dialogue data, the stronger the model will attribute a weight to it. 
Through this process, the graph we built provides not only a more abstract view of (domain, slot, value) tuples for the current dialogue, but also the implicit connections and  transitions in general human dialogues.
This latter part of relations is generally ignored in previous studies. We will show that adding them may provide extra power to the model.

To be more convincing, we show some expected impact of the graph (true examples observed from the test data). For example, domain-domain co-occurrence could indicate that slots of \textit{Hotel} domain more likely need to update if \textit{Train} domain is evoked. Slot-slot co-occurrence could tell that hotels with \textit{stars} usually offer free \textit{parking}. Domain-slot co-occurrence tells whether two (domain, slot) pairs usually co-occur. For example, if (taxi, departure) needs to update value, (taxi, destination) might also need to update.

To effectively leverage such a graph, two key problems should be solved. 
First, we have to aggregate node information of a multi-relational graph. We adopt Relational Graph Convolutional Network (GCN) for this task. 
Second, there are many different slot values (4500 in our datasets) and they are not predefined. It is unrealistic to represent them as nodes in our state graph because this leads to
over-parameterization. 
To address this problem,  value nodes in our graph are merely ``placeholders''. We will not create embeddings for such nodes. Instead, a place is dynamically filled with the corresponding hidden state of the bi-directional Transformer encoder. With these contextualized hidden states and context-independent node and edge embeddings, inference on the dialog state graph incorporates both local context and global information of different domains and slots. Finally, we fuse the graph-based inference with token-level inference of bi-directional self-attention in the encoder. 

The contributions of this work are as follows \footnote{We will release our codes later.}:
\begin{itemize}
\item We propose to build a dialogue state graph for DST which includes elements in the dialogue state and co-occurrence information among them to enhance DST. 

\item We propose a method to deal with such a multi-relational graph with many different values.  

\item Our method (Graph-DST) 
achieves a new state-of-the-art performance on two public datasets (MultiWOZ 2.0 and MultiWOZ 2.1) on DST, showing the usefulness of the state graph.
In particular, by incorporating the state graph, we obtain $1.30\%$ and $1.50\%$ absolute
improvements in joint goal accuracy.
\end{itemize}

\section{Related Work}
\label{sec:related}
Ontology-based DST assumes that not only the domain-slot structure is known, but also the possible values are predefined in an ontology. The goal of DST in this context can be simplified into a value classification/ranking task for each slot \cite{henderson2014word, mrkvsic2017neural, zhong2018global, ren2018towards, ramadan2018large,shan2020contextual}. These studies showed the great impact of ontology to DST. 
A recent work \cite{shan2020contextual} combining ontology and contextual hierarchical attention has achieved high performance on the public datasets MultiWOZ 2.0 and MultiWOZ 2.1. 
In real application situations, however, one cannot always assume that ontology is available. In many cases, slot values are discovered through the conversation rather than predefined (e.g. taxi departure places).  

Open-vocabulary DST addresses this problem: it tries to generate or extract a slot value from the dialogue history  \cite{lei2018sequicity, gao2019dialog, wu2019transferable, ren2019scalable}. In this paper, we focus on open-vocabulary DST.
Some existing approaches encoded the full dialogue history as model input. For example, \citet{wu2019transferable} used copy mechanism to track slot values mentioned anywhere in the dialogue history. 
\citet{le2020non} focused on the efficiency issue. They encoded the slot type information and the dialogue context, then used a non-autoregressive decoder to generate the slot values of the current dialogue state at once. Even though the whole dialogue history is potentially useful, recent work \cite{kim2019efficient, zhu2020efficient} found that utilizing partial dialog history (the most relevant part), combined with previous dialogue state, can achieve higher performance. The reason is that a long dialogue history may contain unimportant elements and it is longer to encode, while a partial dialogue history (the most recent one) may contain the most relevant elements, and it can be effectively complemented by the previous dialogue state.
In general, the partial dialogue history and the previous dialogue state were concatenated as the input to an encoder, which is often based on bi-directional Transformer initialized with BERT. Our method is built on this previous approach, which has proven effective. We supplement it by constructing a dialogue state graph, which provides additional ability to infer the slots to be updated, as well as the update values. The graph encoding is fused with the token-level inference using Transformer.

A work attempting to achieve the same goal as us is \citet{zhu2020efficient} which introduced the Schema graph containing relations among domains and slots. However, the graph was designed to contain only one type of edges and three types of nodes -- domain, slot, and (domain,slot). Only the domain-slot structure was encoded in their graph embeddings. In contrast, our graph additionally encodes three types of co-occurrence between different elements as well as slot values. Besides, the Schema graph is used as prior knowledge shared by all samples in the dataset (aka ontology), while our graph is constructed specifically for each sample. In addition, we also propose a new method to leverage the graph and fuse it with the token-level inference based on dialogue history and previous state.

\section{Method}
\begin{figure*}[t]
\centering
\includegraphics[height=2.6in]{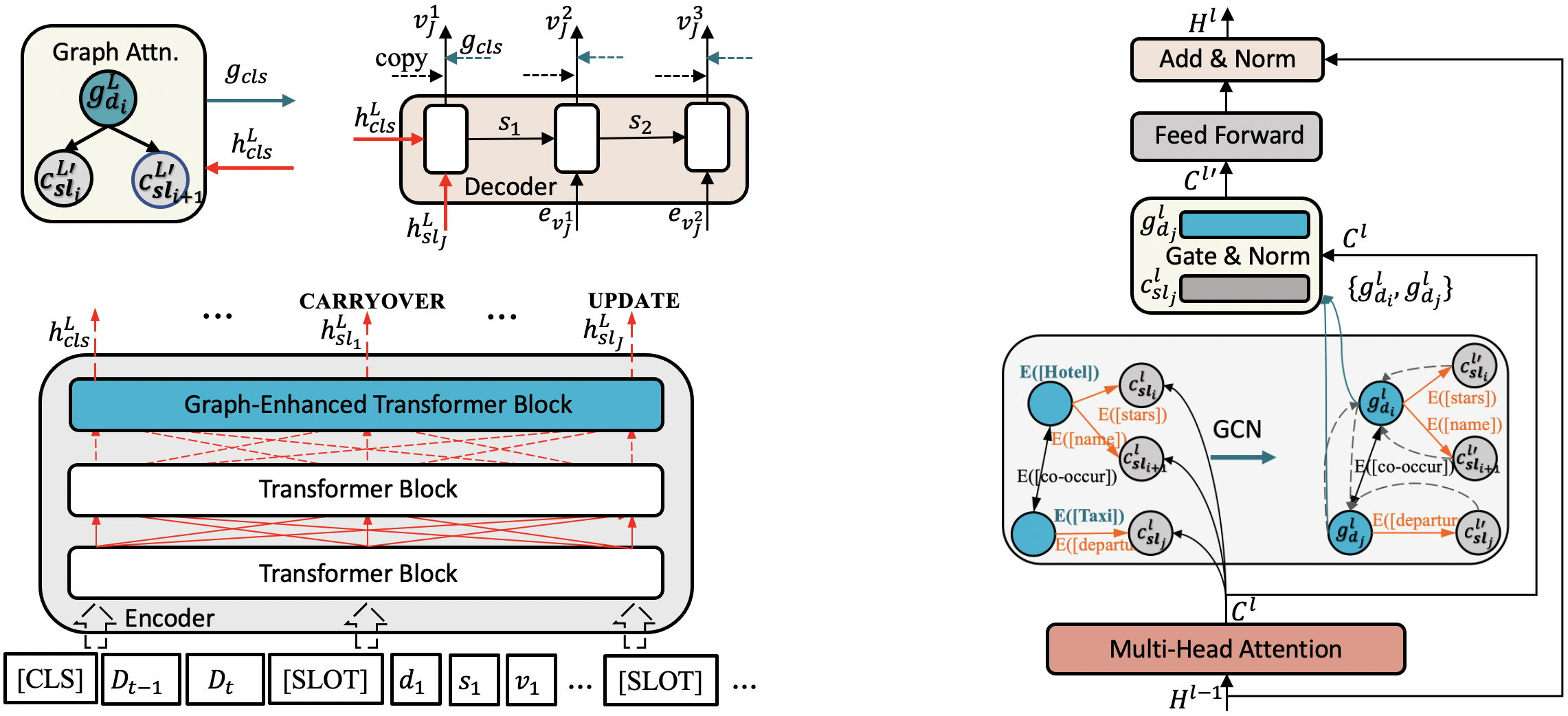}
\caption{(Left) Overview of our predictor-generator framework. The encoder is a bi-directional Transformer initialized with BERT. The decoder generates a value for every slot that has been predicted as UPDATE. (Right) Detailed structure of a graph-enhanced transformer block that is fused with graph inference.}
\label{Fig:model}
\end{figure*}

For multi-domain DST, a conversation with $T$ turns can be represented as ${(D_1, S_1), (D_2, S_2), ..., (D_T, S_{T})}$, where 
$D_t$ is the $t$-th dialog turn consisting of a system utterance and a user response, $S_{t}$ is the corresponding dialogue state. We define $S_{t}$ as a set of ${(d_j, s_j, v_j)|1 \leq j \leq J}$, where $J$ is the total number of (domain, slot) pairs, i.e. $S_{t}$ records slot values of all (domain, slot) pairs. If no information is given about $(d_j, s_j)$, $v_j$ is \textit{NULL}. 
In general, we have a limited number of domains and slots in conversations, but a much larger number of slot values. For instance, in MultiWOZ dataset, there are 5 domains, 17 slots, 30 (domain, slot) pairs, and more than 4500 different values. 

The goal of DST is to predict $S_t$ given $\{(D_1, S_1), ..., (D_{t-1}, S_{t-1}), (D_t)\}$, i.e. we want to extract the state for the current turn $t$ of dialogue, given the previous dialogue history and dialogue states. Following \citet{kim2019efficient}, we do not use the full dialogue history, but only $D_{t-1}$, $D_t$, and $S_{t-1}$ to predict $S_{t}$. The three elements are concatenated as our model input. 

Our work focuses on the utilization of state graph. So we will use a recent effective model proposed by \citet{kim2019efficient} as our base model. In the following sub-section, we will briefly describe the components in the base model before our state graph.

\subsection{Predictor-Generator Framework}
Following \citet{kim2019efficient} and \citet{zhu2020efficient}, our model consists of a state predictor to decide which (domain, slot) pairs need to update values, and a slot value generator. The two components share a bi-directional Transformer encoder initialized with BERT (base, uncased). Figure \ref{Fig:model} (Left) gives an overview of the model. Our training objective is the sum of the slot operation prediction loss and the slot value generation loss. 

\textbf{Encoder} The input is the concatenation of $D_{t-1}$,  $D_t$, and $S_{t-1}$. Each $(d_j, s_j, v_j)$ tuple in $S_{t-1}$ is represented by [SLOT]$\oplus d_j \oplus - \oplus s_j \oplus - \oplus v_j$, where $\oplus$ denotes token concatenation, and [SLOT] and $-$ are separation symbols. Notice that $s_j$ and $v_j$ might consist of several tokens. As illustrated in Figure \ref{Fig:model}, the representations at [SLOT] position $\{\mathbf{h}^L_{sl_j}| 1 \leq j \leq J\}$ are used for state operation prediction. Then, we expect that the hidden states at [SLOT] positions able to aggregate the information from the corresponding $(d, s, v)$ tuples. For example, each $\mathbf{h}^l_{sl_j}$ aggregates information of $(d_j, s_j, v_j)$.

The input representation, i.e. $\mathbf{H}^0$, is the sum of token embedding, position embedding, and type embedding at each position. The encoder is a multi-layer Transformer with bi-directional self-attention, which updates via: $\mathbf{H}^i={\rm Trans}^i(\mathbf{H}^{i-1}), \quad i \in [1,L]$. The last few Transformer layers (blocks) are graph-enhanced, which will be discussed in Section \ref{sec:GTrans}. The outputs of encoder are denoted as $\mathbf{H}^L=[\mathbf{h}^L_{cls}, \mathbf{h}^L_1, ...,\mathbf{h}^L_{sl_1}, ..., \mathbf{h}^L_{sl_J}, ...]$ where $\mathbf{h}^L_{cls}$ is the aggregated representation of the whole layer that summarizes the input as in BERT. It is then used for the query of graph attention and decoder state initialization.

\textbf{State operation predictor} Following \citet{gao2019dialog} and \citet{kim2019efficient}, we use four discrete state operations: CARRYOVER, DELETE, DONTCARE, and UPDATE. State operation predictor, a MLP layer, performs operation classification for each [SLOT] $\{\mathbf{h}^L_{sl_j}| 1 \leq j \leq J\}$.  Specifically, CARRYOVER means to keep the slot value unchanged. DELETE changes the value to NULL, and DONTCARE changes the value to DONTCARE, which means that the slot neither needs to be tracked nor considered important at this turn \cite{wu2019transferable}. Only if UPDATE is predicted that the decoder generates a new slot value for the (domain, slot) pair.

\textbf{Decoder} aims to predict the update value for a slot. It uses a Gated Recurrent Unit (GRU) layer \cite{cho2014properties} with copy mechanism on the input tokens. $\mathbf{h}^L_{cls}$ is used as decoder initial state. If $(d_j, s_j, v_j)$ is predicted to UPDATE value, $\mathbf{h}^L_{sl_j}$ is then fed into the GRU as the initial word representation. 
To obtain the probability distribution over the vocabulary at the $i$-th decoding step, we apply the soft-gated copy mechanism \cite{see2017get}: 
\begin{equation}
\mathbf{P}^i_{s} = {\rm softmax}(\mathbf{W}_e\mathbf{s}_i)
\end{equation}
\begin{equation}
\mathbf{P}^i_{c} = {\rm softmax}(\mathbf{H}^L\mathbf{s}_i)
\end{equation}
where $\mathbf{W}_e \in \mathbb{R}^{|V| \times d_h}$ is the word embedding matrix shared across the encoder and the decoder and $\mathbf{H}^L \in \mathbb{R}^{|x| \times d_h }$ is the encoder outputs. Notice that $\mathbf{P}^i_{s} \in \mathbb{R}^{|V|}, \mathbf{P}^i_{c} \in \mathbb{R}^{|x|}$ and thus we transform $\mathbf{P}^i_{c}$ into the corresponding probability on the vocabulary, denoted as $\mathbf{P'}^i_{c}$. 
Then, the two are combined by:
\begin{equation}
\mathbf{P}^i = \alpha\mathbf{P}^i_{s} + (1-\alpha)\mathbf{P'}^i_{c}
\end{equation}
where $\alpha$ is a scalar value computed via:
\begin{equation}
\alpha = {\rm softmax}(\mathbf{W}_{\alpha}[\mathbf{s}_i; \mathbf{e}_{v^i_j}; \mathbf{c}_i; \mathbf{g}_{cls}])
\end{equation}
where the context vector $\mathbf{c}_i = \mathbf{P}^i_{c}\mathbf{H}^L$ and $\mathbf{g}_{cls}$ is the graph representation from our state graph (see next sub-sections) determined using the aggregated layer representation $\mathbf{h}^L_{cls}$ as the query to attend to the dialog state graph:
\begin{equation}
\mathbf{g}_{cls} = {\rm softmax}(\frac{\mathbf{h}_{cls} \mathbf{G}'^{T}}{\sqrt{d_h}})\mathbf{G}'
\end{equation}
where $\mathbf{G}'$ is the GCN updated graph node embeddings including the hidden states in value placeholders. The graph node embedding is described in the next sub-sections.

\subsection{Dialog State Graph Construction}

The core component of a Transformer Block is the multi-head self-attention mechanism. The bi-directional self-attention could be viewed as a graph neural network whose input is a fully-connected graph. The graph nodes are $\mathbf{H}^{l-1}=[\mathbf{h}^{l-1}_{cls}, \mathbf{h}^{l-1}_1, ...,\mathbf{h}^{l-1}_{sl_1}, ..., \mathbf{h}^{l-1}_{sl_J}, ...]$ (for $l$-th Block). This graph is extremely large with many nodes and edges of the same type. 
The encoder of this graph can learn token-level inference on the combination of dialog history $D_{t-1}, D_t$ and dialog state $S_t$. As we mentioned earlier, this inference is extremely noisy, which may connect unrelated nodes.

Since dialog state is an compact representation of history, we propose to build from it a multi-relational graph (see  Figure \ref{Fig:GDST_intro}) as follows:
Given the previous dialog state $S_{t-1}=\{(d_j, s_j, v_j)|1 \leq j \leq J\}$, each $(d_j, s_j, v_j)$ tuple is represented by the domain node $d_j$ and the value node (``placeholder'') connected by the slot edge. Notice that only (domain, slot) pairs that have filled values (except NULL and DONTCARE) are included in the graph. Furthermore, a co-occurrence edge is added to connect two domain nodes if they co-occur in $S_{t-1}$. The co-occurrence edge is created to capture naturally related domains and transitions in conversations.
Slot edges are uni-directional, while co-occurrence edges are bi-directional. Each domain node or slot edge has its own embedding vector, denoted as E([...]) as in Figure \ref{Fig:model}(Right). Previous works usually assumed (domain, slot) pairs independent from each other. In our work, the domain node and slot edge embeddings trained from data can encode domain-domain, slot-slot, and domain-slot co-occurrences, which may help DST. 
Furthermore, since there are many slot values (4500 in the MultiWOZ dataset) and we do not assume them be pre-defined, each value node in the graph is merely a ``placeholder''. It is not part of graph node embeddings. Instead, these places are dynamically filled with corresponding self-attention outputs at [SLOT] positions, i.e. $\{c^l_{sl_1}, ..., c^l_{sl_J}\}$. With these contextualized hidden states and context-independent graph node and edge embeddings, inference on the dialog state graph can incorporate both local context and global information of different domains and slots.

\subsection{Graph-Enhanced Transformer}
\label{sec:GTrans}
Given a state graph, we need to determine 1) how we update graph representations to combine local context and global domain and slot information; 2) how we fuse graph-based inference with token-level inference of the self-attention mechanism. 

As shown in Figure \ref{Fig:model} (Right), The outputs of $(l-1)$-th Transformer Block $\mathbf{H}^{l-1}=[\mathbf{h}_{cls}^{l-1},  ...,\mathbf{h}_{sl_J}^{l-1}, ...]$  is fed into a multi-head self-attention layer, and the outputs, contextualized representations $\mathbf{C}^{l}=[\mathbf{c}_{cls}^l, ...,\mathbf{c}_{sl_J}^l, ...]$, are computed via: 
\begin{equation}
\mathbf{C}^l = {\rm Concat}(\mathbf{head}_1, ..., \mathbf{head}_h)
\end{equation}
\begin{equation}
\mathbf{head}_j = {\rm softmax}(\frac{\mathbf{Q}_j\mathbf{K}_j^{T}}{\sqrt{d_k}})\mathbf{V}_j
\end{equation}
where $\mathbf{Q}_j, \mathbf{K}_j, \mathbf{V}_j \in \mathbb{R}^{n\times d_k}$ are obtained by transforming $\mathbf{H}^{l-1} \in \mathbb{R}^{n\times d_h}$ using $\mathbf{W}_{j}^{Q}, \mathbf{W}_j^{K}, \mathbf{W}_j^{V} \in \mathbb{R}^{d_h\times d_k}$ respectively. For a value placeholder in the graph, we use the contextualized representation at the corresponding [SLOT] position $c^l_{sl_j}$ 
to fill the placeholder. This representation 
aggregates the information of $(d_j, s_j, v_j)$. Then, we apply multi-relational GCN \cite{vashishth2019composition} to update node representations via pooling features of their adjacent nodes. For example, for a domain node $d$: 
\begin{align*}
\label{eq:gcn}
\mathbf{g}^l_{d} &= f(\mathbf{W}_S (\mathbf{e}_d - \mathbf{e}_S) + \sum_{(v,sl)\in N(d)}\mathbf{W}_O(\mathbf{c}^l_v - \mathbf{e}_{sl}) \\
& + \sum_{d'\in N(d)}(\mathbf{W}_I+\mathbf{W}_O)(\mathbf{e}_{d'} - \mathbf{e}_{co})) \tag{8}\\
\end{align*}
where $(v,sl) \in N(d)$ denotes the neighboring value placeholders and corresponding slot edges and $d'\in N(d)$ denotes the neighboring domain nodes; $\mathbf{c}^l_v$ is the contextualized representation in the placeholder; $\mathbf{W}_S, \mathbf{W}_I, \mathbf{W}_O \in \mathbb{R}^{d_h \times d_h}$ are weight matrices for self-loop edges, ingoing, and outgoing edges respectively; $e_d, e_S, e_{sl}, e_{co} \in \mathbb{R}^{d_h}$ are embeddings for a domain, self-loop edge, a slot edge, and domain co-occurrence edge respectively; and $f$ is an activation function. In Equation \ref{eq:gcn}, an embedding vector instead of a weight matrix is used to represent a type of edge, thereby avoiding the over-parameterization issue of some relational GCNs.

As a result, a updated domain node representation $\mathbf{g}^l_{d_j}$ encodes context and graph structure information. Then, $\mathbf{g}^l_{d_j}$ is fused with the contextualized representations of corresponding [SLOT] positions by:  
\begin{align*}
\mathbf{c}^{l'}_{sl_j} = {\rm LayerNorm}&( \beta^l_{sl_j}\mathbf{c}^l_{sl_j} + (1-\beta^l_{sl_j})\mathbf{g}^l_{d_j}) \tag{9} \\
\beta^l_{sl_j} &= \sigma(\mathbf{W}_{\beta}\mathbf{c}^l_{sl_j}+\mathbf{b}_{\beta}) \tag{10}
\end{align*}
In the examples of Figure \ref{Fig:model} (Right), $\mathbf{g}^l_{d_j}$ is fused with $\mathbf{c}^l_{sl_j}$, and $\mathbf{g}^l_{d_i}$ is respectively fused with $\mathbf{c}^l_{sl_i}$ and $\mathbf{c}^l_{sl_{i+1}}$.

\section{Experiments}

\begin{table*}[t]
\centering
\small
\begin{tabular}{l|lc|cc}
\hline 
\hline
 & \textbf{Model} & \textbf{BERT used} & \textbf{MultiWOZ 2.0} & \textbf{MultiWOZ 2.1}\\
\hline 
 & HJST \cite{eric2019multiwoz} &  & 38.40 & 35.55 \\
& FJST \cite{eric2019multiwoz} & & 40.20 & 38.00 \\
predefined& SUMBT \cite{lee2019sumbt}  & $\surd$ & 42.40 & - \\
ontology & HyST \cite{goel2019hyst} & & 42.33 & 38.10 \\
 & DS-DST \cite{zhang2019find} & $\surd$ & - & 51.21 \\
& DST-Picklist \cite{zhang2019find} & $\surd$ & - &  53.30 \\
& DSTQA \cite{zhou2019multi} & & 51.44 &  51.17 \\
& CHAN-DST \cite{shan2020contextual} & $\surd$ & \textbf{52.68} & \textbf{58.55} \\
\hline
& DST-Span \cite{zhang2019find} & $\surd$ & - & 40.39 \\
& DST-Reader \cite{gao2019jointly} &  & 39.41 & 36.40 \\
& TRADE \cite{wu2019transferable} & & 48.60 & 45.60 \\
open- & COMER \cite{ren2019scalable} & $\surd$ & 48.79 & - \\
vocabulary & NADST \cite{le2020non} & &  50.52 & 49.04 \\
& SAS \cite{hu2020sas} & & 51.03 &  - \\ 
& SOM-DST \cite{kim2019efficient} & $\surd$ & 51.72 &  53.01 \\
& CSFN-DST \cite{zhu2020efficient} & $\surd$ & 51.57 &  52.88 \\
& Graph-DST (ours) & $\surd$ & \textbf{52.78} & \textbf{53.85} \\
& \quad -w/o Graph & $\surd$ & 51.48 & 52.35 \\
\hline
\end{tabular}
\caption{\label{tab:joint} Joint goal accuracy (\%) on the test set of MultiWOZ. Results for the baselines are taken from their original papers. }
\end{table*}

\subsection{Datasets}
We use two publicly available datasets MultiWOZ 2.0 \cite{budzianowski2018multiwoz} and MultiWOZ 2.1 \cite{eric2019multiwoz} in our experiments. MultiWOZ 2.1 is a corrected version of MultiWOZ 2.0. We use the script provided by \citet{wu2019transferable} and \citet{kim2019efficient} to preprocess the datasets, which retain only five domains (restaurant, train, hotel, taxi, and attraction). 
The final test datasets contain 5 domains, 17 slots, 30 (domain, slot) pairs, and more than 4500 different values. Appendix \ref{app: data} gives more statistics of the datasets. 

\subsection{Implementation Details}
Our model is implemented based on the open-source code of SOM-DST \cite{kim2019efficient} \footnote{https://github.com/clovaai/som-dst}. 
We follow their experimental settings. For example, we use greedy decoding for slot value generator. Different learning rate schemes are applied for the encoder, the decoder, and the GCN. We set the learning rate and warmup proportion to 4e-5 and 0.1 for the encoder, and 1e-4 and 0.1 for the decoder. For the GCN, the learning rate is 1e-3, and no warmup is applied. We use a batch size of 32. Our encoder is initialized with BERT (base, uncased). We use one layer GCN to update graph embeddings. The last layer of the encoder is graph-enhanced. The model is trained on two P100 GPU devices for 30 epochs. In the inference, we use the previously predicted dialogue state as input instead of the ground-truth.

\subsection{Baselines}
We compare the performance of our model Graph-DST with both ontology-based models and open vocabulary-based models. 

\textbf{FJST} \cite{eric2019multiwoz} uses a bi-directional LSTM to encode the dialogue history and a feed-forward network to choose the value of each slot.

\textbf{HJST} \cite{eric2019multiwoz} encodes the dialogue history using an LSTM like FJST but utilizes a hierarchical network.

\textbf{SUMBT} \cite{lee2019sumbt} uses BERT to initialize the encoder whose input is the dialogue history and slot-value pairs. Then, it scores each candidate slot-value pair using a non-parametric distance measure.

\textbf{HyST} \cite{goel2019hyst} utilizes a hierarchical RNN encoder and a hybrid approach to incorporate both ontology-based and open vocabulary-based settings.

\textbf{DST-Reader} \cite{gao2019dialog} formulates the problem of DST as an extractive question answering task -- it uses BERT contextualized word embeddings and extracts slot values from the input by predicting spans.

\textbf{DST-Span} \cite{zhang2019find} applies BERT as the encoder and then uses a similar method to DST-Reader.

\textbf{DS-DST} \cite{zhang2019find} uses two BERT-base encoders and designs a hybrid approach for ontology-based DST and open vocabulary DST. It defines picklist-based slots for classification similarly to SUMBT and span-based slots for span extraction as DST Reader.

\textbf{DST-Picklist} \cite{zhang2019find} uses a similar architecture to DS-DST, but it performs only predefined ontology-based DST by considering all slots as picklist-based slots.

\textbf{DSTQA} \cite{zhou2019multi} formulates DST as a question answering problem -- it generates a question asking for the value of each (domain, slot) pair. It heavily relies on a predefined ontology.

\textbf{CHAN-DST} \cite{shan2020contextual} employs a contextual hierarchical attention network based on BERT and uses an adaptive objective to alleviate the slot imbalance problem by dynamically adjust weights of slots during training.

\textbf{TRADE} \cite{wu2019transferable} encodes the full dialogue history using a bi-directional GRU and decodes the value for each (domain,slot) pair using a copy-based GRU decoder.

\textbf{COMER} \cite{ren2019scalable} uses BERT-large as a feature extractor and a hierarchical LSTM decoder whose target sequence is the current  state.

\textbf{NADST} \cite{le2020non} uses a transformer-based non-autoregressive decoder to generate the current state.

\textbf{SAS} \cite{hu2020sas} proposes Slot
Attention and Slot Information Sharing (SAS)
to reduce redundant information’s interference
and improve long dialogue context tracking.

\textbf{SOM-DST} \cite{kim2019efficient} applies predictor-generator framework. The encoder initialized with BERT encodes the previous and current dialogue turns and the dialog state. 

\textbf{CSFN-DST} \cite{zhu2020efficient} introduces the Schema Graph considering relations among domains and slots. Predictor-generator framework is applied. BERT is utilized to initialize a Transformer-based encoder that fuses the context and the schema graph. 

\subsection{Experimental Results}
We report the joint goal accuracy of our model and the baselines on MultiWOZ 2.0 and MultiWOZ 2.1 in Table \ref{tab:joint}. Joint goal accuracy measures whether all slot values predicted at a turn exactly match the ground truth values. The accuracy of baseline models is taken from their original papers.

As shown in the table, our Graph-DST model achieves the highest joint goal accuracy among open-vocabulary DST: $52.78\%$ on MultiWOZ 2.0 and  $53.85\%$ on MultiWOZ 2.1. The only approach that produces higher accuracy than Graph-DST is CHAN-DST (on MultiWOZ 2.1 only). This latter uses ontology, thus benefits from the additional prior knowledge, which is not used in our case. The overall observation is that our Graph-DST can produce state-of-the-art performance compared to the existing works.

In the following sub-sections, we will analyze the impact of using our state graph. Appendix \ref{app:ablation} gives analysis for our model variants. Appendix \ref{app:output} shows some sample outputs of our model. 

\subsection{Graph Effectiveness}
\label{sec:analysis_graph}

Appendix \ref{app:gra_stat} gives the statistics of the dialog state graphs on MultiWOZ 2.1. 
The graph is on average in small scale -- 5 value nodes in 1.7 domains with in total 6 edges. In the train set, only $47\%$ graphs have more than two domain nodes and only $16\%$ graphs have more than three domain nodes. The corresponding values are $52\%$ and $18\%$ in the test set. Furthermore, previous dialog state does not usually contain the ground-truth slot values 
-- only $9.11\%$ training samples and $8.88\%$ test samples. Notice that only for samples associated to more than one domain, their dialog states likely help current DST task. 
We suspect that if more instances have more than one domain nodes in the graph, our method could further show its advantages.

In Table \ref{tab:joint}, we also report our model without state graph. We can see that by incorporating state graph, Graph-DST has been boosted by $1.30\%$ in joint goal accuracy on MultiWOZ 2.0 and $1.50\%$ on MultiWOZ 2.1. This result clearly shows the positive impact of incorporating a state graph, and confirms our assumption that such a state graph can provide complementary information to the implicit graph underlying transformer's self-attention. It is also worth noting that Graph-DST without graph is equivalent to SOM-DST, which uses the same basic architecture (without graph). We can see that our Graph-DST without graph can reproduce similar results. 

We mentioned that Schema graph is a simpler graph than ours, which is constructed on tokens. It is used in CSFN-SDT. The impact of Schema graph has been investigated in \citet{zhu2020efficient}, which showed an improvement of $0.42\%$. This is a much smaller impact than our state graph.
This result is in line with our assumption that more complex domain-slot-value graph incorporating co-occurrence relations can capture more useful information than a token-level graph.

\subsection{Efficiency Analysis}
\label{sec:effi}

\begin{table}[h]
\centering
\begin{tabular}{lll}
\hline 
\hline
\textbf{Model} & \textbf{Joint Accuracy} & \textbf{Latency} \\
\hline 
TRADE & 45.60 & 455ms \\
NADST & 49.04 &  35ms \\
SOM-DST & 53.01 & 48ms \\
Graph-DST (ours) & 53.85 & 52ms \\
\hline
\end{tabular}
\caption{\label{tab:effi} Average inference time per dialogue turn on MultiWOZ 2.1 test set.}
\end{table}

State graph may require extra computation time to process. We show in Table \ref{tab:effi} the latency of our method and some typical models measured on P100 GPU with a batch size of 1. Graph-DST applies the same predictor-generator framework as SOM-DST. Comparing to it, incorporating dialog state graph only adds a small fraction of time. So, adding state graph does not affect efficiency. Graph-DST is about 7.8 times faster than TRADE that generates the values of all the (domain, slot) pairs at every turn of dialogue, and only 0.49 times slower than NADST that uses non-autoregressive decoding. Although NADST shows the highest efficiency, its effectiveness is the worst among the models listed in the table. Overall, the gain in accuracy is worth the slight increase in time. More comparison on Inference Time Complexity (ITC) \cite{ren2019scalable} of Graph-DST and baseline models is provided in Appendix \ref{app:itc}.

\subsection{Error Analysis}
\label{sec:error}

\begin{table}[t]
\small
\centering
\begin{tabular}{lccccc}
\hline 
\hline
\textbf{Model} & \textbf{Attr.} & \textbf{Hotel} & \textbf{Rest.} & \textbf{Taxi} & \textbf{Train} \\
\hline 
Graph-DST & 68.06 & \textbf{51.16} & 64.43 & 57.32 & \textbf{73.82} \\
w/o Graph & 68.19 & 48.11 & 65.65 & 58.41 & 70.46 \\
\hline
\end{tabular}
\caption{\label{tab:domain_jacc}Domain-specific joint accuracy on MultiWOZ 2.1.}
\end{table}

\begin{figure}[t]
\centering
\includegraphics[height=2.2in]{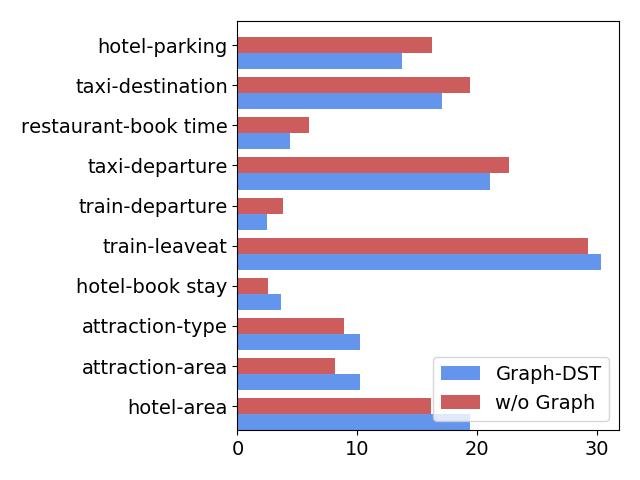}
\caption{Domain-slot \textbf{error} rate (\%) of joint goal accuracy (top-5 and last-5) on test set of MultiWOZ 2.1. }
\label{Fig:ds_error}
\end{figure} 

We analyze the source of errors by comparing with w/o Graph baseline. Table \ref{tab:domain_jacc} shows the domain-specific joint goal accuracy. 
The results indicate that incorporating dialog state graph substantially benefits \textit{Hotel} ($+3.05\%$) and \textit{Train} ($+3.36\%$) domains. 
From the statistics about domains (Appendix \ref{app: data}), we can see that \textit{Hotel} has the largest number of slot types ($=10$), and many slots are related. For example, hotels with \textit{stars} usually have free \textit{parking}. The \textit{Train} domain always co-occurs with another domain. Since our graph embeddings encode domain-domain, domain-slot and slot-slot co-occurrences, Graph-DST can leverage them for these two domain. 

However, Graph-DST performs worse than w/o Graph baseline in the \textit{Restaurant} and \textit{Taxi} domains. As listed in Appendix \ref{app: data}, the slots of \textit{Restaurant} domain are not correlated, such as food, area, name, book day. Besides, models usually need to refer to the current dialogue turn instead of previous dialog state to extract the values for these slots. For \textit{Taxi} domain, considering (domain, slot)-specific results in Figure \ref{Fig:ds_error}, our approach loses advantages only in terms of \textit{leave at}(time) and \textit{arrive by}(time) slots. These two slots also usually need to refer to the current dialogue turn to extract values instead of based on previous dialogue state.
In our approach, the current turn and the previous dialogue state are simply concatenated as input, the model may not know how to extract the needed value from it. We may need to better structure the input so that the dialogue history and the previous state can be combined depending on their usefulness for the expected type of value. We leave this to our future work.

\section{Conclusion}
The existing approaches to DST in open-vocabulary setting exploited the dialogue history and previous dialogue state using transformer. Tokens in them are connected through self-attention. This is equivalent to encoding a large graph. However, noise is easy to generate because of the huge number of connections in such a graph.

In this paper, we proposed to construct another better structured state graph from the previous state, in which domains, slots and values are connected, as well as co-occurrence relations between domains. By using GCN encoding, elements in the graph can be connected, capturing naturally the transitions from a topic to another in conversations. Our experiments on MultiWOZ datasets show the effectiveness of our model. It produced the highest joint goal accuracy in the open-vocabulary DST. Even compared with ontology-based DST, our method is superior to most of them. This confirms the usefulness of our state graph.

Our approach can be further improved in the future by including more types of useful connection in our graph. It would also be interesting to test other approaches to graph encoding.

\bibliography{anthology,acl2020}
\bibliographystyle{acl_natbib}

\clearpage

\appendix

\section{Dataset Statistics}
\label{app: data}
MultiWOZ 2.1 is a refined version of MultiWOZ 2.0 in which the annotation errors are corrected. Some statistics of MultiWOZ 2.1 are reported here.  
\begin{table*}[t]
\centering
\begin{tabular}{llccc}
\hline 
\hline
\textbf{Domain} & \textbf{Slots} & \textbf{Train} & \textbf{Valid} & \textbf{Test} \\
\hline 
Attraction & area, name, type & 8,073 & 1,220 & 1,256 \\
Hotel & price range, type, parking, book stay, book day, book people, & 14,793 & 1,781 & 1,756 \\
& area, stars, internet, name &  &  & \\
Restaurant & food, price range, area, name, book time, book day, book people & 15,367 & 1,708 & 1,726 \\
Taxi & leave at, destination, departure, arrive by & 4,618 & 690 & 654 \\
Train & destination, day, departure, arrive by, book people, leave at & 12,133 &  1,972 & 1,976 \\
\hline
\end{tabular}
\caption{\label{tab:stat} Data statistics of MultiWOZ 2.1 including domain and slot types and number of turns in train, valid, and test set. }
\end{table*}

\begin{table}[h]
\centering
\begin{tabular}{llll}
\hline 
\hline
\multicolumn{3}{c}{Domain Transition} \\
\cline{1-3}
First & Second & Third & Count \\
\hline 
restaurant & \textbf{train} &  - & 87 \\
attraction & \textbf{train} & - & 80 \\
hotel & -  & - & 71 \\
\textbf{train} & attraction & - & 71 \\
\textbf{train} & hotel & - & 70 \\
restaurant &  - & - & 64 \\
\textbf{train} & restaurant & - &  62 \\
hotel & \textbf{train} &  - &  57 \\
\textbf{taxi} & -  & - &  51 \\
attraction & restaurant &  -  &  38 \\
restaurant &  attraction & \textbf{taxi} & 35 \\
restaurant &  attraction & - & 31 \\
\textbf{train} & - & - & 31 \\
hotel & attraction & - & 27 \\
restaurant & hotel & - &  27 \\
restaurant & hotel & \textbf{taxi} & 26 \\
attraction &  hotel & \textbf{taxi} & 24 \\
attraction & restaurant & \textbf{taxi} & 23 \\
hotel &  restaurant &  - &  22 \\
attraction &  hotel & - & 20 \\
hotel & attraction & \textbf{taxi} & 16  \\
hotel & restaurant & \textbf{taxi} & 10  \\
\hline
\end{tabular}
\caption{\label{tab:domain_transit} Statistics of domain transitions that correspond to more than 10 dialogues in the \textbf{test} set of MultiWOZ 2.1. \textit{Train} domain always co-occurrs with another domain. \textit{Taxi} always co-occurrs with another two domains.}
\end{table}

\clearpage

\section{Graph Statistics}
\label{app:gra_stat}
Table \ref{tab:graph_stat} gives the statistics of the dialog state graph. The statistics indicate the graph is on average in small scale. Besides, samples in test set are generally more difficult to handle than samples in training set. 

\begin{table}[h]
\centering
\begin{tabular}{lccc}
\hline 
\hline

 & Train & Valid & Test \\
\hline 
\# edges & 5.56 & 5.77 & 5.83 \\
\# edge types & 5.19 & 5.41 & 5.47 \\
\# nodes & 6.51 & 6.72 & 6.78 \\
\# domains & 1.65 & 1.73 & 1.73\\
\# values & 4.86 & 4.99 & 5.05 \\
$\geq2$ domains & 47\% & 53\% & 52\% \\
$\geq3$ domains & 16\% & 17\% & 18\% \\
in dialog state & 9.11\% & 9.39\% & 8.88\% \\
\hline
\end{tabular}
\caption{\label{tab:graph_stat} Statistics of the dialog state graph nodes and edges on MultiWOZ 2.1. ``in dialog state'' gives the percentage of the case when the previous dialog state contains the ground-truth slot values of current DST task.}
\end{table}

\section{Inference Time Complexity (ITC)}
\label{app:itc}
\begin{table}[h]
\centering
\begin{tabular}{lcc}
\hline 
\hline
\multirow{2}{*}{Model}  & \multicolumn{2}{c}{Inference Time Complexity} \\
\cline{2-3}
 & Best & Worst \\
\hline 
SUMBT & $\Omega(JM)$ & $O(JM)$ \\
DS-DST & $\Omega(J)$ & $O(JM)$ \\
DST-picklist & $\Omega(JM)$ & $O(JM)$  \\
DST Reader & $\Omega(1)$ & $O(J)$ \\
TRADE & $\Omega(J)$ & $O(J)$ \\
COMER &  $\Omega(1)$ & $O(J)$ \\
NADST &  $\Omega(1)$ & $O(1)$ \\
ML-BST &  $\Omega(J)$ & $O(J)$ \\
SOM-DST &  $\Omega(1)$ & $O(J)$ \\
CSFN-DST &  $\Omega(1)$ & $O(J)$ \\
Graph-DST (ours) & $\Omega(1)$ & $O(J)$ \\
\hline
\end{tabular}
\caption{\label{tab:itc} Inference Time Complexity (ITC) of our method and baseline models. We report the ITC in both the best case and the worst case for more precise comparison. $J$ indicates the number of slots, and $M$ indicates the number of values of a slot.}
\end{table}

\clearpage

\section{Model Variant Analysis}
\label{app:ablation}
We tested several variants of our model and the experimental results are listed in Table \ref{tab:variants}. 

\begin{table}[h]
\centering
\begin{tabular}{lcc}
\hline 
\hline
\textbf{Model} & \textbf{Joint Acc.} & \textbf{Slot Acc.}\\
\hline 
w/o Graph & 52.35 & 97.31 \\  
Graph-DST & 53.85 & 97.39 \\   
\hline
+2nd gcn & 52.93 & 97.36 \\  
+2nd GTrans & 52.35 & 97.31 \\ 
G-attn with $\mathbf{h}^L_{sl}$ & 52.85 & 97.35 \\

\hline
\end{tabular}
\caption{\label{tab:variants} Joint goal accuracy and slot operation prediction accuracy on  MultiWOZ 2.1.}
\end{table}

\textbf{+2nd GTrans} denotes applying graph to enhance two last Transformer blocks. This weakens model performance probably because it brings a larger change on the pre-trained BERT, which may not be beneficial given the small number of training data (about 55K training samples).

\textbf{+2nd gcn} represents applying two GCN layers, which enables a node access to its neighbors' neighbors. However, it is known that multi-layer GCN overly smooths graph representations. As listed in Appendix \ref{app:gra_stat}, there are only 1.7 domain nodes on average, and $\sim 50\%$ graphs have only one domain node, in which case 2 GCN layers do not expand the neighborhood of a domain node but largely smooth the representations. 

\textbf{G-attn with $\mathbf{h}_{sl}^L$} uses $\mathbf{h}^L_{sl_j}$ that aggregates ($d_j$, $s_j$, $v_j$) information instead of $\mathbf{h}^L_{cls}$ that summarizes the encoder input as in BERT as the query for the graph attention. However, experimental result shows that using $\mathbf{h}^L_{cls}$ is better.

\section{Sample Prediction Output}
\label{app:output}

\begin{figure*}[t]
\centering
\includegraphics[height=5.6in]{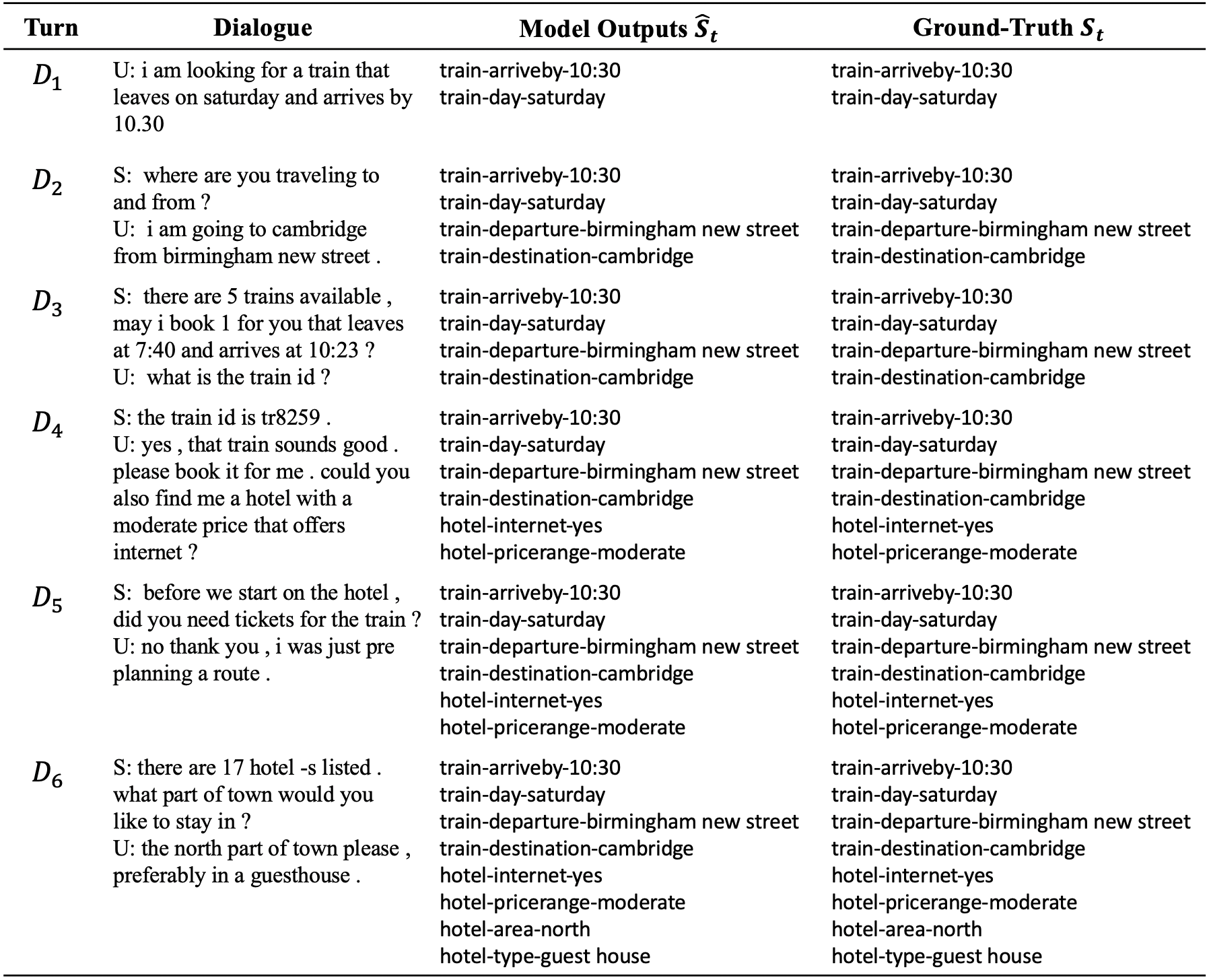}
\caption{The outputs of Graph-DST in a dialogue (index: MUL0671) in the test set of MultiWoz 2.1. To save space, we omit NULL slot values from the figure.}
\label{Fig:case_p1}
\end{figure*} 

\begin{figure*}[t]
\centering
\includegraphics[height=6in]{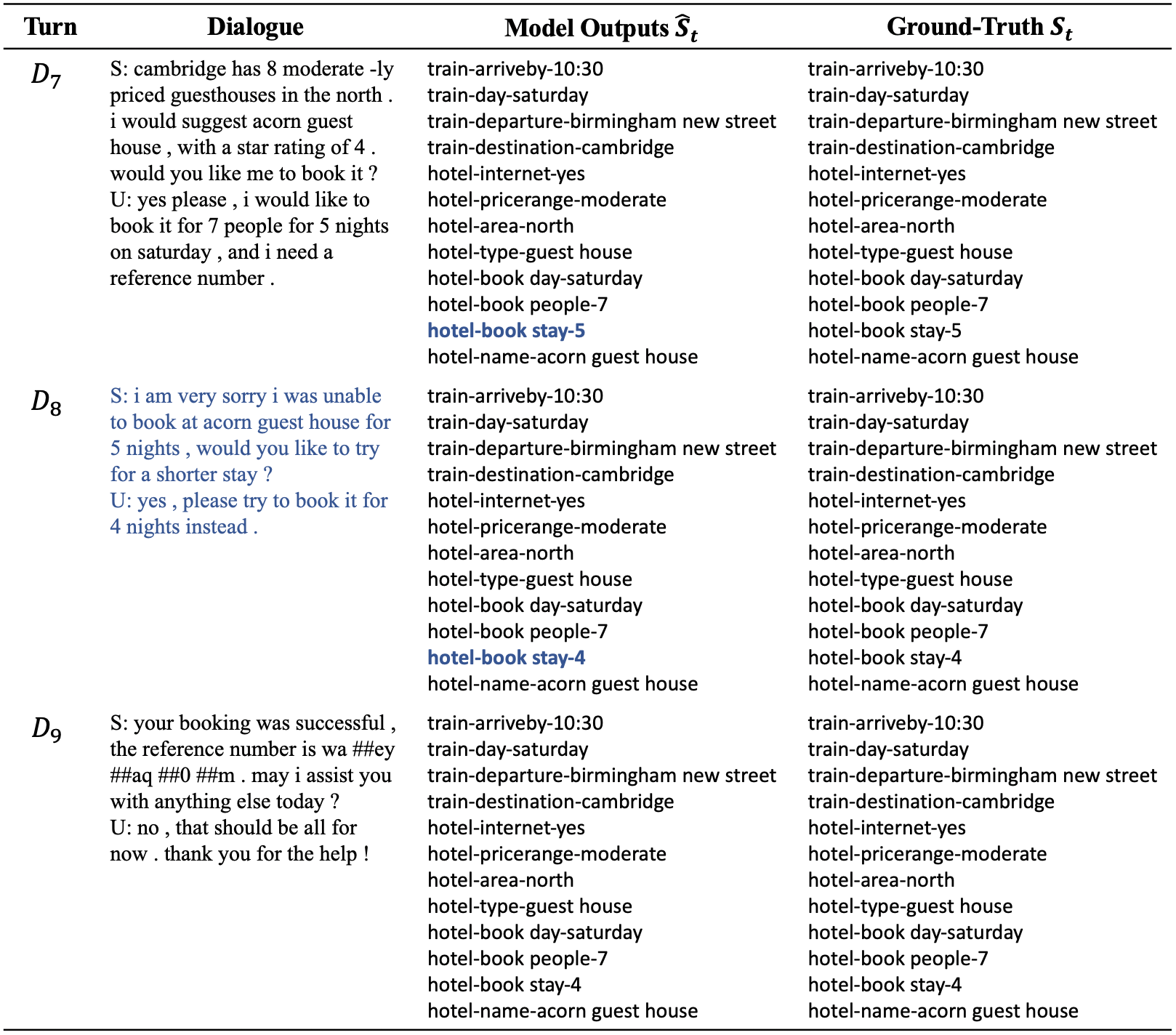}
\caption{The outputs of Graph-DST in a dialogue (index: MUL0671) in the test set of MultiWoz 2.1. To save space, we omit NULL slot values from the figure.}
\label{Fig:case_p1}
\end{figure*}

\end{document}